\documentclass[conference]{IEEEtran}
\IEEEoverridecommandlockouts
\usepackage{cite}
\usepackage{amsmath,amssymb,amsfonts}
\usepackage{algorithmic}
\usepackage{graphicx}
\usepackage{textcomp}
\usepackage{multicol, multirow}
\usepackage{xcolor}
\usepackage{url}
\def\BibTeX{{\rm B\kern-.05em{\sc i\kern-.025em b}\kern-.08em
    T\kern-.1667em\lower.7ex\hbox{E}\kern-.125emX}}
\begin{document}

\title{A Federated Learning Benchmark on Tabular Data: 
Comparing Tree-Based Models and Neural Networks}

\author{\IEEEauthorblockN{1\textsuperscript{st} William Lindskog}
\IEEEauthorblockA{\textit{Corporate Research and Development} \\
\textit{DENSO Automotive Deutschland GmbH}\\
Munich, Germany \\
w.lindskog@eu.denso.com}
\and
\IEEEauthorblockN{2\textsuperscript{nd} Christian Prehofer}
\IEEEauthorblockA{\textit{Corporate Research and Development} \\
\textit{DENSO Automotive Deutschland GmbH}\\
Munich, Germany \\
c.prehofer@eu.denso.com}}

\maketitle

\begin{abstract}
Federated Learning (FL) has lately gained traction as it addresses how machine learning models train on distributed datasets. FL was designed for parametric models, namely Deep Neural Networks (DNNs).Thus, it has shown promise on image and text tasks. However, FL for tabular data has received little attention. Tree-Based Models (TBMs) have been considered to perform better on tabular data and they are starting to see FL integrations. In this study, we benchmark federated TBMs and DNNs for horizontal FL, with varying data partitions, on 10 well-known tabular datasets. Our novel benchmark results indicates that current federated boosted TBMs perform better than federated DNNs in different data partitions. Furthermore, a federated XGBoost outperforms all other models. Lastly, we find that federated TBMs perform better than federated parametric models, even when increasing the number of clients significantly. 
\end{abstract}

\section{Introduction}\label{sec:introduction}
With data regulations e.g. GDPR \cite{albrecht2016gdpr}, one faces challenges when centralizing distributed datasets. Federated Learning (FL) has gained traction as it advocates privacy-preserving Machine Learning (ML). In FL, a copy of a global model is sent from a server to \textit{clients} who train the copy on local data and send back updated parameters. The local model updates are aggregated to form a new (global) model at server-side. 

FL was originally designed for \textit{parametric} models \cite{konevcny2016federated}, e.g. Deep Neural Networks (DNNs), in which the relationship between input and output data is defined by a mathematical function. FL starts to see real-world applications using text \cite{ banabilah2022federated} and images \cite{doshi2022federated}. FL for tabular data has received less attention. Tabular data are commonly organized in tables, formed by rows and columns, for which columns usually represent features of observed data. Tabular data are common and often found in industry and academia. For this type of data, \textit{non-parametric} models, mainly Tree-Based Models (TBMs) e.g. decision trees or random forests \cite{kern2019tree} have been successful in centralized settings. 
\begin{figure}[tb]
    \centering
    \includegraphics[width=\columnwidth]{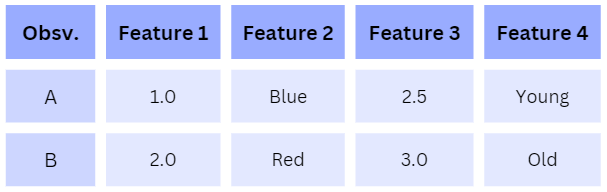}
    \caption{Tabular data, each row is a unique observation and the columns indicate features. Values can be numerical and categorical.}
    \label{fig:tabular_data}
\end{figure}
Recently, federated TBMs have been constructed, federated random forests \cite{liu2020federated}, federated gradient boosted decision trees \cite{fedtree}, and federated XGBoost \cite{tian2020federboost}. It is however still uncertain whether these federated TBMs are better than federated parametric models on tabular data. To our knowledge, this has not been investigated and we seek to answer this open question. 

A challenge in FL is data heterogeneity in the distributed datasets. Commonly assumed, data are non-independent and identically distributed (non-IID) and can have consequences on FL model convergence. Few benchmarks have studied FL models when aggregating client results in various IID settings \cite{li2022federated} but often separating federated parametric models and TBMs, or not including TBMs for tabular data. It is observed that federated TBMs have received little attention in comparison with federated parametric models, however research in federated TBMs is gaining traction. 

In this study, we focus on horizontal FL in which distributed datasets share feature space but not sample space. We benchmark open-source federated versions of 3 different TBMs and 3 DNNs on 10 well-known tabular datasets. We include several different data partitioning strategies, label-, feature-, and quantity distribution skew, to study model performance on the datasets. Our main contributions to research are:
\begin{itemize}
    \itemsep0em
    \item We are the first to benchmark federated- TBMs and parametric models on a wide set of tabular datasets. 
    \item In our benchmark, federated boosted tree-based models outperform federated DNNs on tabular data, even in non-IID data settings, especially federated XGBoost. 
    \item We find that federated TBMs perform better than federated parametric models, even when increasing the number of clients significantly. 
\end{itemize}
\section{Related Work}\label{sec:related_work}
Following subsections include models and FL for tabular data and FL for tabular data.
\subsection{Tree-Based Models for Tabular Data}\label{subsec:(RELATED_WORK)_TBModels_for_Tabular_Data}
Researchers have considered TBMs to outperform DNNs on tabular data. Models such as random forests, single decision trees and gradient boosted decision trees (GBDTs) \cite{kern2019tree}, namely XGBoost e.g. (XGB) \cite{chen2015xgboost}, are among the most prominent choices of TBMs from tabular data. Parametric TBMs have also been developed, referred to as \textit{differentiable trees}. Research has designed differentiable TBMs \cite{popov2019neural, kontschieder2015deep, yang2018deep}, and the key point is "smooth" decision functions, using e.g. Entmax \cite{peters2019sparse}, in the internal tree structure. Compared with non-parametric TBMs, this enables a differentiable tree function and routing, which can utilize gradient optimization. However, their performance could be further studied on more datasets, including multi-class classification tasks \cite{popov2019neural}. 

\subsection{Deep Neural Networks}\label{subsec:(RELATED_WORK)_deep_neural_networks}
DNNs \cite{lecun2015deep} have seen many implementations for several types of data e.g. image and text \cite{goodfellow2016deep}. In \cite{gorishniy2021revisiting}, they conclude that DNNs can perform better than TBMs on tabular data in few cases, however, they show no broad sign of superiority. According to them, a ResNet \cite{he2016deep} model can serve as a baseline on tabular data. Another study supports the claim that simple regularized multi-layer perceptron (MLP) can outperform TBMs on tabular data \cite{kadra2021well}. \cite{shwartz2022tabular} showed that tree-based models requires less tuning than DNNs, thus they are preferable to utilize for tabular data. They nevertheless state that a deep ensemble can achieve superior performance. Other research suggest that DNNs' performance on tabular data is inferior to the of tree-based algorithms \cite{grinsztajn2022tree, iman2021comparative, shankar2022tree}. 

Attention-based models \cite{vaswani2017attention, dosovitskiy2020image} have lately sparked much interest and researchers have constructed specific models for tabular data. In \cite{shwartz2022tabular}, they mention TabNet \cite{arik2021tabnet} to be one of the better preforming attention-based models on tabular data. Researchers suggest other attention-based models, namely TabTransformer \cite{huang2020tabtransformer}, AutoInt \cite{song2019autoint}, and FT-Transformer \cite{gorishniy2021revisiting}. 

\subsection{Federated Learning for Tabular Data}\label{subsec:(RELATED_WORK)_Federated_Learning_for_Tabular_Data}
FL was initially designed for parametric models, namely DNNs, and can easily be integrated with models in Subsection \ref{subsec:(RELATED_WORK)_deep_neural_networks}. As mentioned, TBMs can provide superior performance, yet their federated implementations are new, and fewer have been made open-sourced. Today, there are many FL programming packages available: FATE \cite{fate_2022}, Flower \cite{beutel2020flower}, TensorFlow Federated (TFF), PySyft \cite{ziller2021pysyft}, FLUTE \cite{dimitriadis2022flute} and FedML \cite{chaoyanghe2020fedml}, yet few support federated TBMs. Both \cite{yang2019tradeoff} and \cite{tian2020federboost} implemented a federated XGB (F-XGB) for tabular data in horizontal FL (HFL), but tested it only on binary classification and did not open-source their code. There are few active frameworks for F-XGB and most of them require a multi-machine set-up \cite{IBM_xgboost}. \cite{li2020practical} was, to our knowledge, the first to present an open-source framework for GBDTs for HFL, that can run on stand-alone machines, and tested it on several tabular datasets. It is however uncertain how their federated TBMs compare to federated DNNs. Recently, they have extended their work to create Unifed \cite{liu2022unifed} which benchmarks open-source federated frameworks. In this study, tabular data receives little attention. 

\begin{figure}[tb]
    \centering
    \includegraphics[width=\columnwidth]{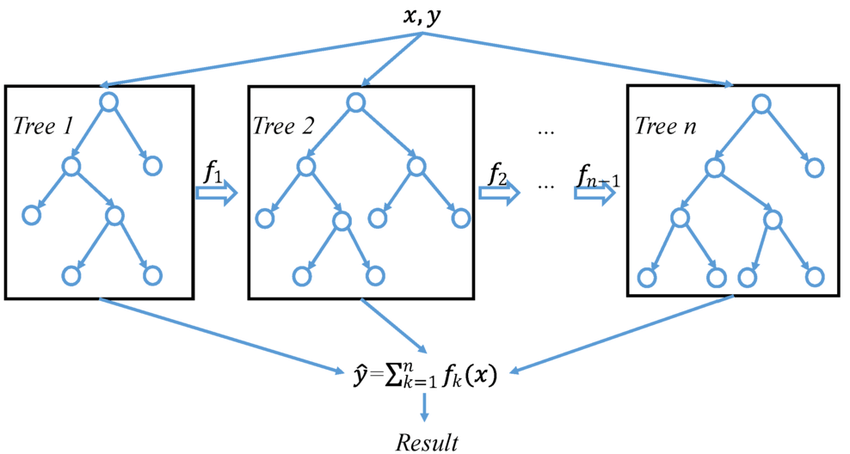}
    \caption{Architecture of XGBoost \cite{wang2019hybrid}. Using so called "weak learners", XGBoost combines predictions these learners to output a final prediction.}
    \label{fig:xgboost}
\end{figure}

One of the main challenges in FL is heterogeneous data in distributed datasets \cite{li2020federated, banabilah2022federated}. This is commonly referred to as non-IID data and researchers have lately developed algorithms that consider this e.g. FedProx \cite{li2020fedprox}, FedNova \cite{wang2021novel}, SCAFFOLD \cite{karimireddy2020scaffold}. These aggregating algorithms have been benchmarked, yet comparing the models have received little attention. The most common FL approach is to utilize federated averaging, or FedAvg \cite{mcmahan2017communication}. The authors claim that FedAvg is robust to non-IID data but their non-IID settings could be extended. Such categories of non-IID data settings are discussed in the work from \cite{zhu2021federated}. 
\section{Benchmark Design}\label{section:benchmark_design}
We source both parametric and non-parametric models for our benchmark study. A baseline option is an MLP, based on the work from \cite{gorishniy2021revisiting, kadra2021well}. 
\cite{gorishniy2021revisiting} state ResNet can outperform certain TBMs on tabular data, thus we include a federated ResNet (F-ResNet). We include a third parametric model, TabNet \cite{arik2021tabnet}, an attention-based model which has already seen federated implementations \cite{lindskog2022federated}. TabNet is the attention-based model that has been included in most benchmark studies \cite{borisov2022deep} for tabular data, thus we include it as \textit{FedTab}. 

We also seek to include the most prominent federated TBMs. FATE has federated ensemble TBMs but their work is based on the work from \cite{cheng2021secureboost}, thus mainly designed for vertical FL. We highlight the work from \cite{li2020federated} which has resulted in FedTree \cite{fedtree}, a fast, open-source package, specifically designed for federated TBMs. Based on \cite{li2020federated, liu2022unifed}, there is reason to select FedTree over FATE if one seeks to simulate federated models on stand-alone machines. We therefore use their federated random forests (F-RFs) and federated GBDTs (F-GBDTs) in our study. Lastly, we use NVIDIA's new framework NVFlare \cite{roth2022nvidia} and construct a federated XGBoost. It supports horizontal FL and we include it as our final model in this benchmark study as F-XGB. 
\subsection{Data Partitioning Strategies}\label{subsec:(BENCHMARK_DESIGN)_Data_Partitioning_Strategies}
Argued by \cite{li2022federated, zhu2021federated}, deploying specific partitioning strategies permits careful control and quantification of non-IID properties. 
Thus, we choose to simulate federated settings using partitioning strategies on tabular datasets. The authors of the studies elaborate on which non-IID data cases can be prevalent in a federated setting. 
\subsubsection{Label Distribution Skew}
 Label distributions $C(y_i)$ can vary among clients. For example, specialized hospitals can have data on different diseases to various degrees. We can break label distribution skewness into 2 parts: 
 \begin{itemize}
     \item \textbf{quantity-based label imbalance} - Clients are assigned a fixed number of labels. \cite{li2022federated} suggest to provide each client with $k$ different label IDs. The samples of each label are distributed to the clients, randomly and equally, with respect to the label IDs they own. Thus, the number of labels a client owns is fixed. We denote this as $\#c = k$.
     \item \textbf{distribution-based label imbalance} - Each client receives a proportion of the samples. Researchers often use Dirichlet distribution as a prior distribution \cite{huang2005maximum}. As \cite{li2022federated}, we also use Dirichlet distribution. We denote our method as $p_k \sim Dir_N(\beta)$ and provide client $C_i$ with a proportion $p_{k,i}$ of data points of class $k$. Parameter $\beta, \beta > 0$ represents concentration, and by choosing to run experiments with a e.g. lower $\beta$, the partition is less balanced. We present this strategy as $p_k \sim Dir(\beta)$. 
 \end{itemize}
\subsubsection{Feature Distribution Skew}
Feature distribution skew suggest that feature distributions $C(x_i)$ differ across clients, even if the knowledge $C(y_i|\mathbf{x_i})$ remains unchanged: 
\begin{itemize}
    \item \textbf{Real-world federated tabular datasets} - There are few open-source federated tabular datasets and they include a natural feature distribution for an arbitrary task. We define federated datasets as data with natural keyed generation process (keys refers to unique users), and (2) distribution skew across users/devices \cite{caldas2018leaf}. 
    \item \textbf{Noise-based feature imbalance} - Advocates adding Gaussian noise to clients' datasets. Randomly and equally divide the entire dataset into $n$ (client) subsets. The feature values are modified by adding noise $\mathbf{\hat{x}}$ $\sim$ $Gau(\sigma \cdot i/N)$ to client $C_i$. One can regulate $\sigma$ to achieve different levels of feature dissimilarity across client data. We present this strategy as $\mathbf{\hat{x}}$ $\sim$ $Gau(\sigma)$
\end{itemize}
\subsubsection{Quantity Skew}
Quantity skew advocates different sizes of client datasets $|D^i|$. The data distribution can still be identical across clients. As for \textit{label distribution skew}, we allocate data to clients according to Dirichlet distribution. A $q_i$ proportion, $q \sim Dir_N(\beta)$ of the dataset is allocated to client $C_i$. Parameter $\beta$ controls the imbalance level of quantity skew. We denote quantity skew partition strategy as $q \sim Dir(\beta)$. 
\subsection{Benchmark Study Settings}\label{subsec:Benchmark_Study_Settings}
In our study, we focus on tabular datasets for regression, binary- and multi-class classification tasks. We also include datasets with varying number of features and classes. We include 10 public datasets: FEMNIST and Synthetic (SYN) \cite{caldas2018leaf}, Insurance Price Prediction (PRICE) \cite{Kaggle_Insurance_dataset}, (ADULT, abalone (ABAL), Agaricus (AGARI) mushrooms, forest type (FOREST), heart disease (HEART), wine quality (WINE)) \cite{Dua:2019} and California house prices (CALI) \cite{pace1997sparse}. We use $20\%$ test data and $80\%$ training data, allocated to the clients. The data partition values ($\sigma$ and $\beta$) are default values from \cite{li2022federated}. 

Flower \cite{beutel2020flower} is used for constructing federated parametric models, FedTree \cite{fedtree} for F-RF and F-GBDT, and NVFlare for F-XGB. We show our hyperparameter search space in Table \ref{tab:(BENCHMARK_STUDY_SETTINGS)_hyperparameters}. For parametric models, Adam \cite{kingma2014adam} is selected as optimizer, loss function is mean squared error and cross-entropy for regression and classification,  respectively. FedTab's (TabNet) other parameters are set to default values \cite{arik2021tabnet}. 

 \begin{table}[tb]
\caption{Statistics of tabular datasets in benchmark study. Datasets with no number of classes are regression tasks. K is thousand.}
\label{tab:(BENCHMARK_STUDY_SETTINGS)}
\vskip 0.15in
\begin{center}
\begin{small}
\begin{sc}
\begin{tabular}{lllll}
\hline
Data & \# Train & \# Test & \# Features & \# Class\\
\hline
ADULT & 39.0k & 9.8k & 14 & 2 \\
AGARI & 6.5k & 1.6k & 22 & 2 \\
FEMNIST & 306k & 76k &784 & 10 \\
FOREST & 92.9k & 23.2k & 54 & 7 \\
HEART & 0.59k & 0.15k & 11 & 2 \\
SYN & 121k & 30k & 30 & 30\\
\hline
ABAL & 3.3k & 0.8k & 8 & --- \\
CALI & 16.3k & 4.1k & 9 & --- \\
PRICE & 1.07k & 0.27k & 6 & --- \\
WINE & 5.2k & 1.3k & 12 & --- \\
\hline
\end{tabular}
\end{sc}
\end{small}
\end{center}
\vskip -0.1in
\end{table}

\begin{table}[tb]
\caption{Hyperparameters and respective values and search space. Column "Type" refers to whether the parameter is specific for a model or general, in this case specified. }
\label{tab:(BENCHMARK_STUDY_SETTINGS)_hyperparameters}
\vskip 0.15in
\begin{center}
\begin{small}
\begin{sc}
\begin{tabular}{llc}
\hline
Type & Parameter & Value\\
\hline
\multirow{5}{*}{General} & FL Strategy & FedAvg\\
& Learning Rate & $\{0.001, 0.005, 0.01\}$\\
& nbr\_clients & $\{3, 5, 10, 15, 25, 50\}$\\
& epochs/trees & 300\\
& early\_stopping & 200\\
\hline
\multirow{3}{*}{MLP/ResNet} & Hidden Layers & 3\\
& Nodes per layer & $\{50, 60, 70, 80, 90\}$\\
& batch\_size & $\{8, 16, 32, 64, 128\}$\\
\hline
\multirow{6}{*}{FedTab} & n\_d & $\{8, 16, 24\}$\\
& n\_a & $\{8, 16, 24\}$\\
& n\_steps & $\{2, 3, 4\}$\\
& gamma & $\{1.2, 1.3, 1.5, 1.8\}$\\
& momentum & $\{0.01, 0.05, 0.1, 0.2\}$\\
& batch\_size & $\{8, 16, 32, 64, 128\}$\\
\hline
TBMs & Depth & $\{3, 4, 5, 6, 7\}$\\
\hline
\end{tabular}
\end{sc}
\end{small}
\end{center}
\vskip -0.1in
\end{table}
\section{Benchmark Results}\label{sec:Benchmark_Results}

\begin{table*}[tb]
\caption{Comparing models on classification and regression datasets, homogeneous setting. Top-1 accuracy and mean squared error, averaged over 5 runs, including standard deviation (\%) for classification and regression tasks respectively, using 10 clients. $\uparrow$ and $\downarrow$ indicate if a scores should be higher or lower. $\star$, values are multiplied with $10^{-9}$. } 
\label{tab:(RESULTS)_Homogeneous}
\vskip 0.15in
\begin{center}
\begin{small}
\begin{sc}
\begin{tabular}{r|c|c|c|c|c|c}
\hline
Data & F-MLP & F-ResNet & FedTab & F-RF & F-GBDT &  F-XGB\\
\hline
\hline
ADULT $\uparrow$& $83.82\pm0.33$ & $84.77\pm0.10$ & $85.02\pm0.20$ & $75.24\pm0.38$& $85.77\pm0.30$& $\mathbf{87.50\pm1.33}$ \\
\hline
AGARICUS $\uparrow$ &$100.00\pm0.00$& $100.00\pm0.00$&$100.00\pm0.00$&$51.91\pm0.08$& $100.00\pm0.00$&$100.00\pm0.00$\\
\hline
FEMNIST $\uparrow$&$90.95\pm0.12$&$91.27\pm0.09$&$92.45\pm0.08$&$91.41\pm0.05$&$92.69\pm0.24$&$\mathbf{93.53\pm0.15}$\\
\hline
FOREST $\uparrow$& $80.64\pm0.31$ & $84.29\pm0.32$ & $80.31\pm0.77$ &$76.19\pm0.12$ &$\mathbf{87.32\pm0.52}$ &$76.50\pm0.60$\\
\hline
HEART $\uparrow$&$84.38\pm0.37$ & $86.92\pm0.32$& $86.09\pm0.60$&$85.65\pm0.70$ &$88.01\pm0.45$& $\mathbf{88.10\pm0.50}$\\
\hline
SYN $\uparrow$ &$81.97\pm0.46$ &$83.52\pm0.40$ & $87.00\pm0.30$&$86.25\pm0.25$ &$86.88\pm0.29$ &$\mathbf{87.19\pm0.21}$ \\
\hline
\hline
ABALONE $\downarrow$ & $4.88\pm0.14$&$4.44\pm0.12$ & $\mathbf{3.97\pm0.25}$& --- & $5.63\pm0.17$& $6.13\pm0.21$ \\
\hline
CALI$^{\star}$ $\downarrow$ & $13.14\pm0.21$ & $7.29\pm0.23$ & $5.04\pm0.76$ & --- & $\mathbf{2.71\pm0.11}$ & $2.97\pm0.39$ \\
\hline
PRICE$^{\star}$ $\downarrow$& $0.052\pm0.005$& $0.030\pm0.003$& $0.033\pm0.004$& --- & $0.021\pm0.003$& $\mathbf{0.016\pm0.002}$\\
\hline
WINE $\downarrow$ & $0.75\pm0.03$ & $0.63\pm0.01$ & $0.66\pm0.02$ & --- & $\mathbf{0.43\pm0.01}$ & $0.54\pm0.05$\\
\hline
\hline
\# Best & 0 & 0 & 1 & 0 & 3 & 5\\
\hline
\end{tabular}
\end{sc}
\end{small}
\end{center}
\vskip -0.1in
\end{table*}

In this section, we firstly present the models' performance in terms of accuracy and mean squared error for homogeneous data distribution. We also present R2 scores for regression tasks in Table \ref{tab:(RESULTS)_r2_scores}. Thereafter, we present model performances for various data partitions. The default metrics are top-1- accuracy and mean squared error for classification and regression respectively. We use 10 clients and show the results in Table \ref{tab:(RESULTS)_Homogeneous}. Moreover, we find the best hyperparameter in our search space (Table \ref{tab:(BENCHMARK_STUDY_SETTINGS)_hyperparameters}) after extensive search. All experiments are executed on a 8GB NVIDIA GeForce RTX 3080 GPU. 

In a homogeneous setting, we find that the federated gradient boosted decision trees (F-GBDT and F-XGB) perform better than the other models. They are performing well on all tasks; regression, binary- and multi-class classification. In most cases, their scores are significantly higher (or lower) than remaining models'. F-XGB is the only model in homogeneous setting to be a "best performer" on all sorts of tasks. Additionally, F-XGB has the best R2 scores in most regression tasks and can explain the variance in the target value well using the input. We see that FedTab often outperforms the other parametric models yet that its standard deviation is relatively high. It is the best performing model on the ABALONE datasets, on which other parametric models seems to do well. F-MLP is however the worst performing model in a homogeneous setting. In some cases, it performs similarly to F-ResNet. F-MLP struggles on regression tasks and is the worst performing model on almost all datasets for regression. Moreover, when adding more participating clients e.g. 15, 25, or 50, we see the same relation in performance between models, only that the overall performance decreases. 
\begin{table}[tb]
\caption{R2 scores in (\%) for various model on regression task. }
\label{tab:(RESULTS)_r2_scores}
\vskip 0.15in
\begin{center}
\begin{small}
\begin{sc}
\begin{tabular}{lcccc}
\hline
Data&F-ResNet&FedTab&F-GBDT&F-XGB\\
\hline
ABAL & $53.1\pm1.8$ & $54.3\pm1.5$& $46.2\pm0.8$ & $40.2\pm2.8$ \\
CALI & $42.1\pm2.2$ & $66.8\pm1.9$ & $77.1\pm1.0$& $79.1\pm2.9$ \\
PRICE & $86.3\pm0.1$ & $87.4\pm0.3$ & $87.3\pm0.3$ & $89.9\pm0.1$ \\
WINE & $36.5\pm1.4$ & $15.4\pm2.9$ &$ 43.1\pm1.1$ & $26.8\pm5.0$ \\
\hline
\# Best & 0 & 1 & 1 & 2\\
\hline
\end{tabular}
\end{sc}
\end{small}
\end{center}
\vskip -0.1in
\end{table}

We notice that F-RF does seem to opt for one value in binary classification tasks, and is completely unable to generalize for regression task, thus we put "-" as value in Table \ref{tab:(RESULTS)_Homogeneous}. For multi-class classification, it is also not the best performing model. Due to these poor results, we do not include it in further experiments. 

\begin{table*}[tb]
\caption{Comparing models on classification datasets, label distribution skew. Top-1 accuracy and standard deviation (\%)  averaged over 5 runs using 10 clients. } 
\label{tab:(RESULTS)_Label_DISTRIBUTION}
\vskip 0.15in
\begin{center}
\begin{small}
\begin{sc}
\begin{tabular}{l|c|c|c|c|c|c}
\hline
Data & Partitioning & F-MLP & F-ResNet & FedTab & F-GBDT &  F-XGB\\
\hline
\hline
ADULT & $p_k \sim Dir(0.5)$ & $78.07\pm2.11$ & $76.81\pm1.50$ & $78.29\pm1.45$& $85.26\pm0.18$& $\mathbf{87.33\pm0.91}$ \\
& $\#c=1$ & $82.40\pm2.24$ & $83.21\pm1.10$& $82.99\pm0.89$& $85.16\pm0.19$ & $\mathbf{86.90\pm0.52}$\\
\hline
AGARICUS & $p_k \sim Dir(0.5)$ & $86.79\pm0.39$ &$99.06\pm0.15$ & $100.00\pm0.00$&$100.00\pm0.00$ & $100.00\pm0.00$\\
& $\#c=1$ & $96.73\pm1.04$& $100.00\pm0.00$& $100.00\pm0.00$&$100.00\pm0.00$ & $100.00\pm0.00$\\
\hline
FEMNIST & $p_k \sim Dir(0.5)$ & $87.99\pm0.25$&$89.88\pm0.24$&$91.98\pm0.30$&$92.22\pm0.26$&$\mathbf{93.02\pm0.22}$\\
& $\#c=1$ & $10.52\pm0.05$ & $11.42\pm0.09$ & $14.09\pm0.40$& $78.10\pm1.20$& $\mathbf{79.92\pm0.98}$\\
& $\#c=2$ & $80.10\pm0.67$ & $59.10\pm2.50$ & $81.41\pm0.60$ & $82.19\pm0.62$& $\mathbf{82.20\pm0.60}$\\
& $\#c=3$ & $80.40\pm0.11$ & $68.40\pm1.21$ & $82.33\pm0.24$& $82.43\pm0.30$& $\mathbf{83.23\pm0.36}$\\
\hline
FOREST & $p_k \sim Dir(0.5)$ & $49.15\pm0.15$&$49.15\pm0.03$ & $50.77\pm0.09$ & $\mathbf{80.18\pm0.18}$ & $78.33\pm0.41$\\
& $\#c=1$ & $14.29\pm0.04$ & $14.31\pm0.04$ & $18.44\pm0.41$ & $\mathbf{69.81\pm0.71}$& $64.69\pm0.35$\\
& $\#c=2$ &$31.72\pm1.25$ &$38.89\pm0.69$ & $42.44\pm0.42$& $\mathbf{75.19\pm0.31}$& $71.11\pm0.22$\\
& $\#c=3$ & $40.46\pm0.30$&$45.91\pm0.33$ & $47.02\pm0.20$& $\mathbf{77.92\pm0.24}$& $77.52\pm0.30$\\
\hline
HEART & $p_k \sim Dir(0.5)$ &$82.50\pm0.40$ & $85.23\pm0.33$& $85.21\pm0.69$&$87.65\pm0.49$& $\mathbf{88.01\pm0.52}$\\
& $\#c=1$ & $81.95\pm0.41$ & $83.02\pm0.60$& $80.49\pm0.83$& $86.55\pm0.50$ & $\mathbf{86.77\pm0.43}$\\
\hline
SYNTHETIC & $p_k \sim Dir(0.5)$ &$81.80\pm0.48$ &$83.33\pm0.41$ & $85.06\pm0.40$&$86.66\pm0.30$ &$\mathbf{86.72\pm0.23}$ \\
& $\#c=1$ &$6.69\pm1.05$& $6.99\pm1.10$& $7.72\pm1.33$& $\mathbf{8.55\pm1.54}$& $7.92\pm1.11$\\
& $\#c=2$ & $15.65\pm0.89$& $15.25\pm0.85$ & $17.22\pm0.90$& $18.22\pm0.75$& $\mathbf{18.49\pm0.71}$\\
& $\#c=3$ & $26.65\pm0.22$&$26.02\pm0.41$ & $28.11\pm0.77$& $32.01\pm0.71$& $\mathbf{34.51\pm0.69}$\\
\hline
\# Best & & 0&0 &0 &5&  11\\
\hline
\end{tabular}
\end{sc}
\end{small}
\end{center}
\vskip -0.1in
\end{table*}

Next, we complete experiments using label distribution skew and we only test models on classification datasets since their labels are discrete and not continuous. We show our results in Table \ref{tab:(RESULTS)_Label_DISTRIBUTION}. We find that federated TBMs outperform federated parametric models for all datasets. Every model experiences a slight decrease in performance but this is expected as task difficulty is higher than for homogeneous setting. F-XGB is the best performing model and is superior in most binary- and multiclass classification tasks. The parametric models experience a sharp decrease in performance for multiclass classification when they each client receives data from few unique classes, especially when $c=1$. Federated TBMs do not experience this sharp decrease in performance. However, the more unique labels the clients are given, the better the performance. This is in line with findings from \cite{li2022federated}. It is clearly showed in the results from Synthetic dataset. It contains 30 classes and when $c=1$, only 10 classes are included in the training data, thus the performance is low for all models. We see however how federated TBMs achieve higher performance for this dataset even though the conditions are challenging. 

We thereafter test both feature- and quantity distribution skew partitioning. For feature distribution skew, we recognize that categorical values should not be included when adding noise to feature values. Thus, we exclude Agaricus Mushroom dataset since all its features are categorical. We present our benchmark results in Table \ref{tab:(RESULTS)_Feature_and_Quantity_Distribution}. In terms of model performance in relation to other models', we see  no major differences to performances in previous tables. F-XGB is the best performing model, followed by F-GBDT. FedTab outperforms other models on the ABALONE dataset. Noticeable, other parametric models perform well on this dataset in comparison with federated TBMs. Nevertheless, federated TBMs show significantly higher performances on other datasets compared with parametric models. Overall, there are slight decreases in performance, with few models seeing improved performance. The best performing model's score is usually worse than in homogeneous setting, which is expected. Nevertheless, it is worth mentioning that no model is significantly affected by either feature- or quantity distribution skew. 

\begin{table*}[tb]
\caption{Comparing models on classification and regression datasets, feature (upper half)- and quantity (lower half) skew. Top-1 accuracy and mean squared error with standard deviation (\%), averaged over 5 runs using 10 clients. $\uparrow$ and $\downarrow$ shows if scores should be higher or lower. $\star$, values are multiplied with $10^{-9}$.}
\label{tab:(RESULTS)_Feature_and_Quantity_Distribution}
\vskip 0.15in
\begin{center}
\begin{small}
\begin{sc}
\begin{tabular}{r|c|c|c|c|c|c}
\hline
Data & Partitioning & F-MLP & F-ResNet & FedTab & F-GBDT &  F-XGB\\
\hline
ADULT $\uparrow$& \multirow{10}{*}{$\mathbf{\hat{x}} \sim Gau(0.1)$}&$83.80\pm0.28$& $84.70\pm0.11$& $85.29\pm0.29$& $85.81\pm0.30$& $\mathbf{87.59\pm1.51}$\\
FEMNIST $\uparrow$&&$90.72\pm0.14$&$91.20\pm0.10$&$92.48\pm0.20$&$92.63\pm0.25$&$\mathbf{93.53\pm0.18}$\\
FOREST $\uparrow$& &$80.58\pm0.38$& $84.39\pm0.14$& $80.38\pm0.77$& $\mathbf{87.15\pm0.48}$& $76.75\pm0.69$\\
HEART $\uparrow$&&$84.18\pm0.35$ & $86.90\pm0.30$& $85.80\pm0.63$&$87.69\pm0.42$& $\mathbf{88.02\pm0.52}$\\
SYN $\uparrow$ && $81.91\pm0.50$ &$83.11\pm0.62$ & $86.84\pm0.50$&$86.89\pm0.22$ &$\mathbf{87.02\pm0.16}$ \\
& \\
ABALONE $\downarrow$& &$5.03\pm0.15$& $4.47\pm0.18$& $\mathbf{3.98\pm0.21}$& $5.71\pm0.22$& $6.28\pm0.30$\\
CALI$^{\star}$ $\downarrow$& & $13.14\pm0.30$ & $7.22\pm0.26$ & $5.04\pm0.71$ & $2.95\pm0.22$ & $\mathbf{2.92\pm0.29}$ \\
PRICE$^{\star}$ $\downarrow$& &$0.054\pm0.005$& $0.032\pm0.003$& $0.033\pm0.004$ & $0.022\pm0.003$& $\mathbf{0.017\pm0.002}$\\
WINE $\downarrow$& & $0.72\pm0.03$ & $0.66\pm0.02$ & $0.67\pm0.04$  & $\mathbf{0.48\pm0.02}$ & $0.57\pm0.03$\\
\hline
\# Best & & 0 & 0& 1 & 2 & 6\\
\hline
ADULT $\uparrow$&\multirow{11}{*}{$q \sim Dir(0.5)$} &$83.87\pm0.31$& $84.78\pm0.10$& $85.25\pm0.25$& $85.80\pm0.25$& $\mathbf{87.56\pm1.40}$\\
AGARICUS $\uparrow$& &$100.00\pm0.00$& $100.00\pm0.00$& $100.00\pm0.00$& $100.00\pm0.00$& $100.00\pm0.00$\\
FEMNIST $\uparrow$&&$90.74\pm0.15$&$91.25\pm0.08$&$92.60\pm0.30$&$92.69\pm0.21$&$\mathbf{93.54\pm0.18}$\\
FOREST $\uparrow$& &$80.56\pm0.35$& $84.34\pm0.12$& $80.40\pm0.79$& $\mathbf{87.35\pm0.50}$& $76.56\pm0.61$\\
HEART $\uparrow$&&$84.18\pm0.35$ & $86.90\pm0.30$& $85.80\pm0.63$&$87.69\pm0.42$& $\mathbf{88.02\pm0.52}$\\
SYN $\uparrow$ && $81.93\pm0.47$ &$83.35\pm0.41$ & $86.82\pm0.55$&$86.90\pm0.25$ &$\mathbf{87.09\pm0.18}$ \\
\\
ABALONE $\downarrow$& &$4.85\pm0.17$& $4.48\pm0.12$& $\mathbf{4.01\pm0.29}$& $5.75\pm0.25$& $6.16\pm0.24$\\
CALI$^{\star}$ $\downarrow$& & $13.00\pm0.25$ & $7.31\pm0.24$ & $4.98\pm0.80$ & $\mathbf{2.75\pm0.19}$ & $3.07\pm0.31$ \\
PRICE$^{\star}$ $\downarrow$& &$0.053\pm0.005$& $0.032\pm0.003$& $0.032\pm0.004$ & $0.021\pm0.003$& $\mathbf{0.016\pm0.002}$\\
WINE $\downarrow$& & $0.74\pm0.03$ & $0.65\pm0.01$ & $0.65\pm0.01$  & $\mathbf{0.44\pm0.02}$ & $0.54\pm0.04$\\
\hline
\# Best & & 0 & 0& 1 & 3 & 5\\
\hline
\end{tabular}
\end{sc}
\end{small}
\end{center}
\vskip -0.1in
\end{table*}


\subsection{Varying Number of Clients}
\begin{figure}[tb]
\vskip 0.2in
\begin{center}
\centerline{\includegraphics[width=\columnwidth]{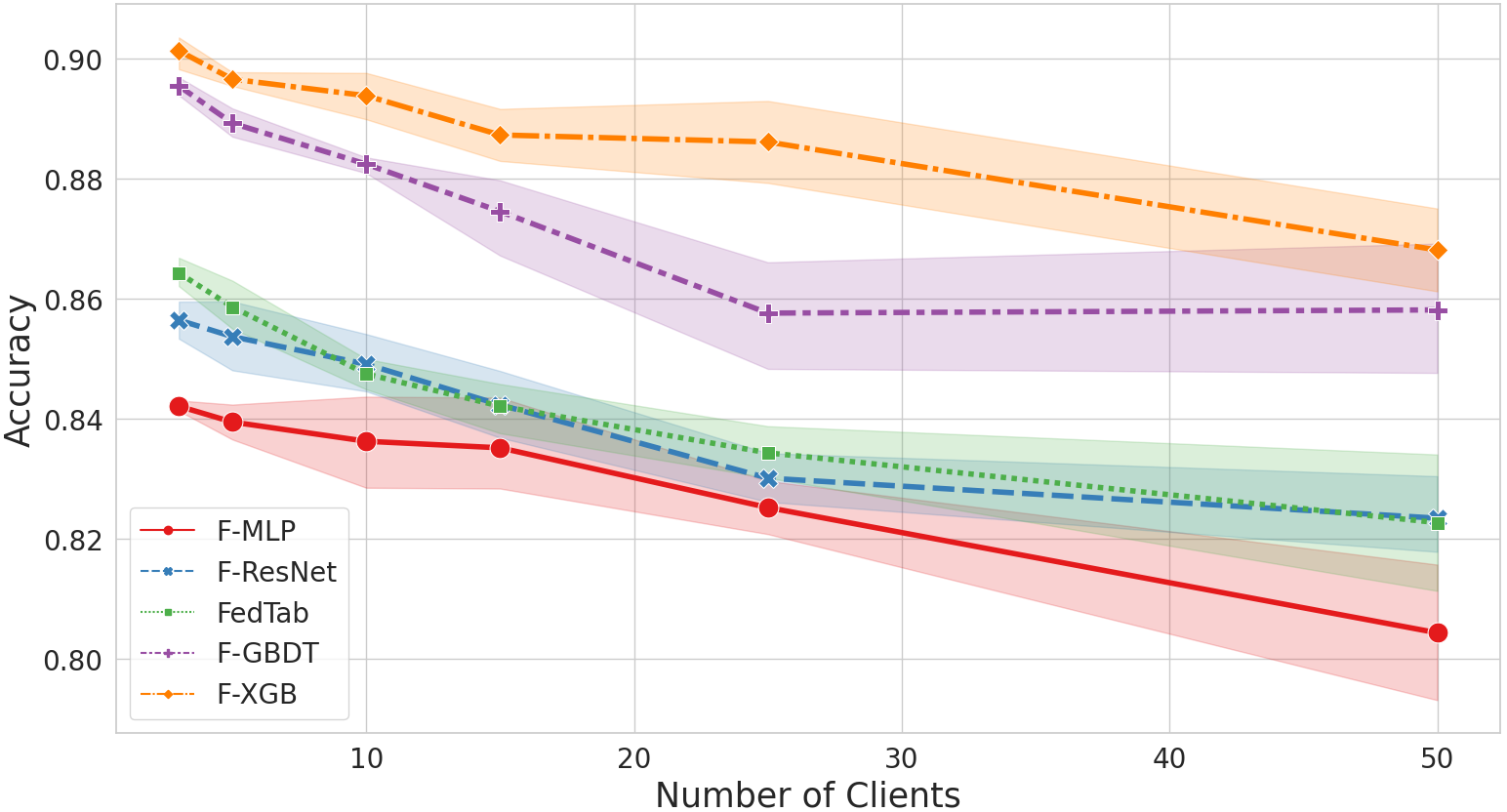}}
\caption{Models' test accuracy for 3, 5, 10, 15, 25 and 50 clients on Heart Disease dataset in homogeneous setting.}
\label{fig:client_scores_homo}
\end{center}
\vskip -0.2in
\end{figure}

\begin{figure}[tb]
\vskip 0.2in
\begin{center}
\centerline{\includegraphics[width=\columnwidth]{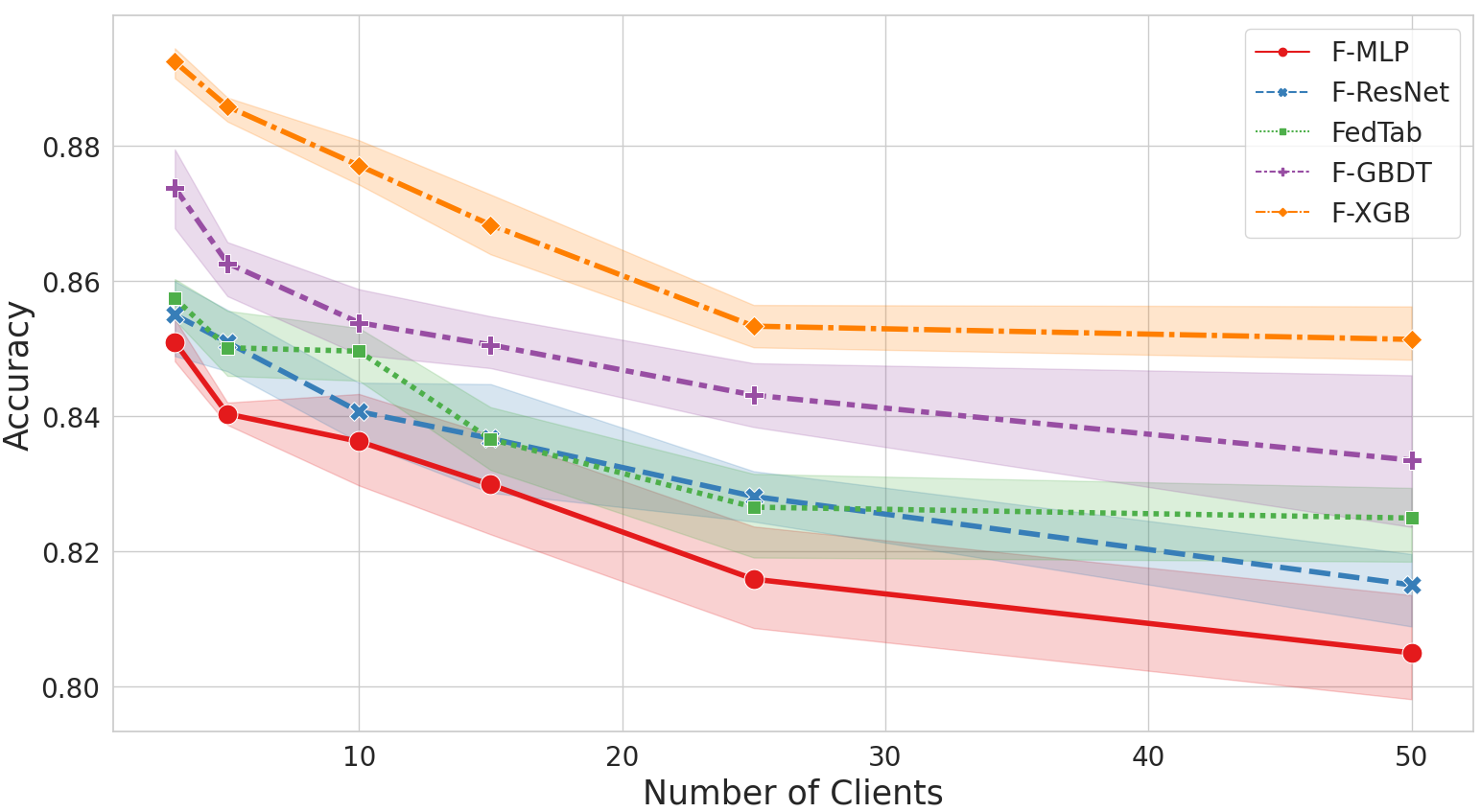}}
\caption{Models' test accuracy for 3, 5, 10, 15, 25 and 50 clients on Adult dataset in homogeneous setting.}
\label{fig:homo_adutl}
\end{center}
\vskip -0.2in
\end{figure}

\begin{figure}[tb]
\vskip 0.2in
\begin{center}
\centerline{\includegraphics[width=\columnwidth]{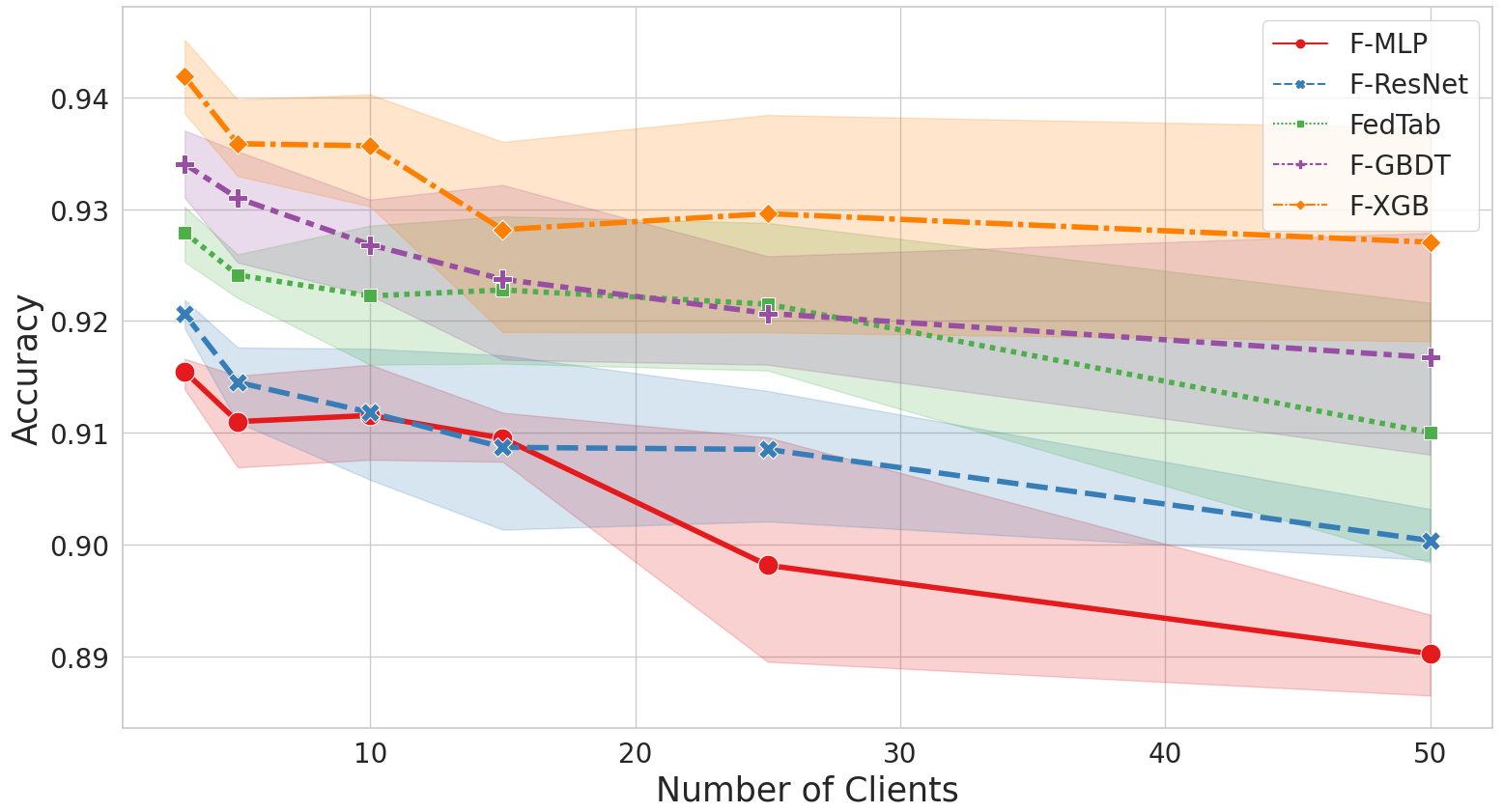}}
\caption{Models' test accuracy for 3, 5, 10, 15, 25 and 50 clients on FEMNIST dataset in homogeneous setting.}
\label{fig:client_scores_homo_fem}
\end{center}
\vskip -0.2in
\end{figure}

\begin{figure}[tb]
\vskip 0.2in
\begin{center}
\centerline{\includegraphics[width=\columnwidth]{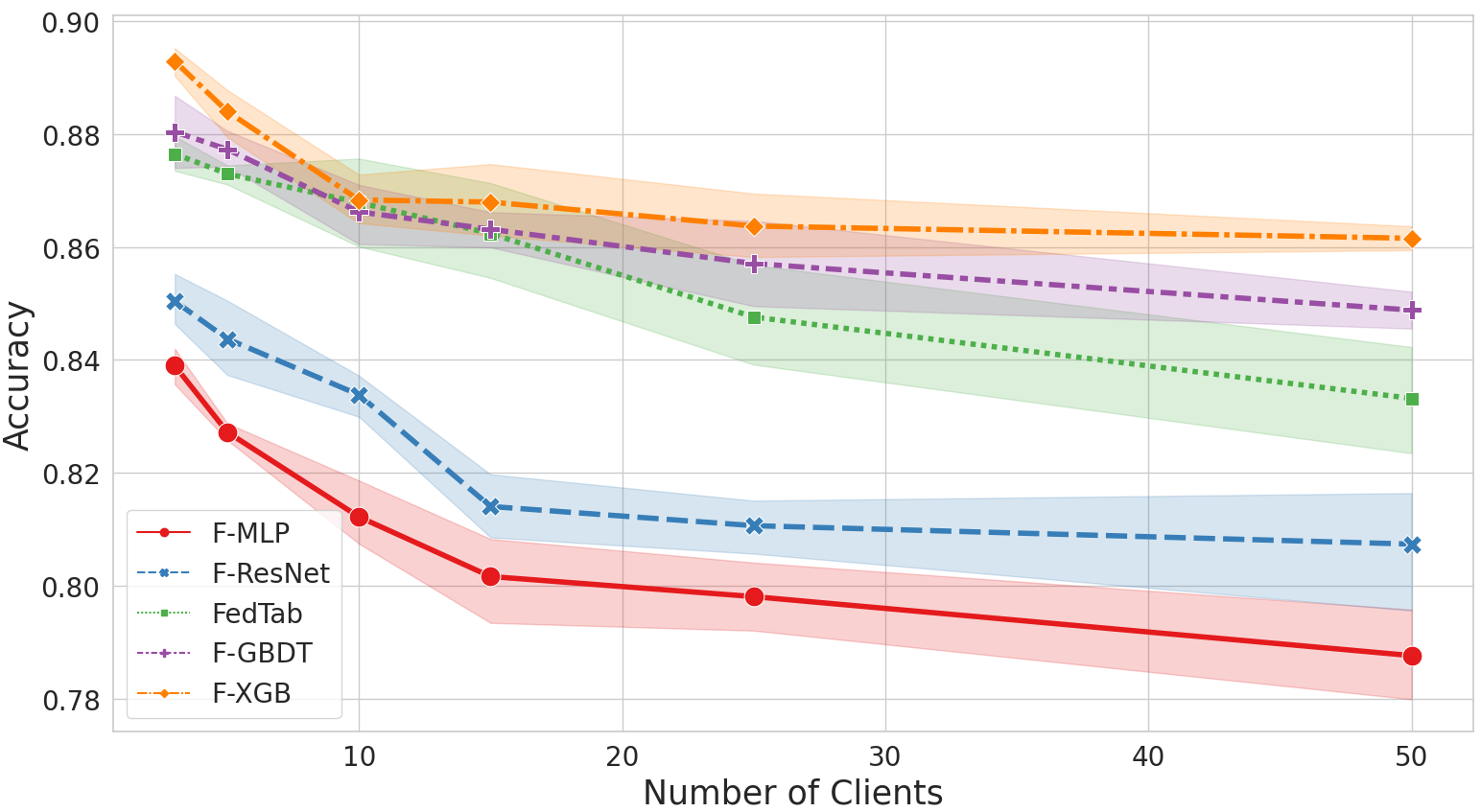}}
\caption{Models' test accuracy for 3, 5, 10, 15, 25 and 50 clients on Synthetic dataset in homogeneous setting.}
\label{fig:homo_syn}
\end{center}
\vskip -0.2in
\end{figure}

Having studied the partitions skews using 10 participating clients, we evaluate how model performance scales with an increasing number of clients in a homogeneous setting. We show performance for increasing number of clients for the Heart Disease, Adult, FEMNIST, and Synthetic datasets in Figure \ref{fig:client_scores_homo}, \ref{fig:homo_adutl}, \ref{fig:client_scores_homo_fem}, and \ref{fig:homo_syn} respectively. We include a significant interval in our plots to show how model performance may deviate from mean values, based on 5 runs. 

We identify a clear trend, performance decreases as number of participating clients increases in a homogeneous setting. We shall also mention that this is found for the other datasets as well. F-XGB is still the best performing model, even when including 50 clients. In a homogeneous setting, the standard deviation increases when the number of participating increases. This is seen in all figures. We also recognize that there is an evident performance gap between federated parametric models and federated TBMs, even as the number of client increases. This is clearly illustrated in Figure \ref{fig:client_scores_homo}. FedTab is still outperforming the other parametric models but we should highlight that its training time is significantly longer than that of F-MLP and F-ResNet. The binary datasets, Heart Disease and Adult, are quite small datasets and we see a significant drop in performance as number of clients increases. This prominent drop in performance is not necessarily observed when scaling number of clients on the larger datasets, FEMNIST and Synthetic. F-XGB's performance only drops with approximately 1.5\% units between 3 $\rightarrow$ 50 clients.

We see similar behaviour when applying our models and scaling the number of clients on the other classification datasets. For the regression tasks, we see a slight increase in mean squared error when increasing the number of clients. Also there, F-XGB is triumphant and performs the best overall.  
\section{Discussion}\label{sec:(DISCUSSION)}
Our insights derived from the benchmark results are hereby discussed and help us define directions that future research in FL might take. First, our benchmark results suggest that federated TBMs are better than federated parametric models on tabular data. In our benchmark, we include real-world federated tabular datasets and tabular datasets to which we apply various partitions. We recognize that F-XGB is best performing model, outperforming models on classification and regression tasks. We show how it scales in a homogeneous setting when adding more clients to training rounds. Federated TBMs perform better than federated parametric models even when the number of clients increase from 3 $\rightarrow$ 50. This can be seen in Figures \ref{fig:client_scores_homo}, \ref{fig:homo_adutl}, \ref{fig:client_scores_homo_fem}, and \ref{fig:homo_syn}. FedTab is the best performing federated parametric model but it shall be stated that it takes long time to train in comparison with all other models. We elaborate that the superiority of federated TBMs might be rooted in their ability to better disregard uninformative features. This was put forward by \cite{grinsztajn2022tree} and could have similar effects in a federated setting. Researchers should thus examine whether the findings from \cite{grinsztajn2022tree} holds in a federated setting.

\subsection{Federated Tree-Based Models} 
As seen in our benchmark results, federated TBMs are robust to various data partitions and relatively fast. 

Questions did however arise when noticing that federated random forest did not perform well, see Table \ref{tab:(RESULTS)_Homogeneous}. Random forests are in a centralized setting a competitive model with much inert explainability. However, its federated implementation seems to suffer on tabular data, even though there are publications with their respective federated random forest implementations \cite{de2020dfedforest, liu2020federated}. Thus, researchers should focus on designing federated forests that can compete with both federated versions of DNNs and GBDTs, and benchmark them on many relevant datasets. 

Moreover, as we have seen, federated TBMs can perform well in terms of commonly logged scores e.g. accuracy, R2. Researchers should also study other important metrics such as compute utilization and running time, communication effort and explainability. TBMs are intuitive models and can provide much insights about feature importance and data selection by analyzing the tree structures. With regards to explainability, this is an emerging topic but mainly for centralized learning. 
\subsection{Partition Robust Federated Models}
As in Table \ref{tab:(RESULTS)_Label_DISTRIBUTION}, no federated model was able to efficiently handle label distribution skew for tabular data. Partition robust FL models should be developed to handle such cases, in which local optima can be far from global optima. This can include more robust federated model architectures, and / or in combination with other algorithms. Inspiration might be derived from the field of personalized federated learning \cite{tan2022towards}, in which non-IID data setting is common. 

Illustrated in Figures \ref{fig:client_scores_homo}, \ref{fig:homo_adutl}, \ref{fig:client_scores_homo_fem}, and \ref{fig:homo_syn}, the performance of the models differ as the number of clients increase. Evidently, federated TBMs provide better performance than federated parametric models in a homogeneous setting. Researchers should examine whether this holds in other partitions e.g. label skew. A problem can arise when clients are given data from 1 unique label and the number of clients is less than unique number of labels. This can be further studied. 

\section{Conclusion}\label{sec:Conclusion}
Research in FL is seeing a growing interest and researchers have begun to develop FL for tabular data. Tabular data are common data in many industrial and academic datasets, and FL should support this type of data. FL has mainly focused on integration with DNNs and federated TBMs have received little attention but their performance on tabular data is strong. 

To address the above research gap, we benchmark open-source federated tree based models and DNNs on 10 well-known tabular datasets. Our benchmark includes model performance on different non-IID data partitions. For the non-IID data partitions, we evaluate label, feature, and quantity distribution skew. We have achieved several key contributions. Firstly, we introduced a comprehensive benchmarking approach, evaluating federated TBMs and parametric models across tabular datasets. This innovation sheds light on model performance in various contexts.

Our findings reveal that federated boosted TBMs, especially federated XGBoost, outperform federated DNNs for tabular data, particularly in non-IID data scenarios. Moreover, our research highlights the consistent superiority of federated TBMs over federated parametric models, even with a substantial increase in the number of clients. These findings underscore the robustness and effectiveness of TBMs in federated learning environments. Overall, our work advances the understanding of model performance and scalability in this critical field.

\section*{Acknowledgement}
This work was partly supported by the TRANSACT project. TRANSACT (https://transact-ecsel.eu/) has received funding from the Electronic Component Systems for European Leadership Joint Under-taking under grant agreement no.  101007260.  This joint undertaking receives support from the European Union’s Horizon 2020 research and innovation programme and Austria,  Belgium,  Denmark,  Finland,  Germany, Poland, Netherlands, Norway, and Spain.

\bibliography{main.bib}
\bibliographystyle{ieeetr}

\end{document}